\documentclass[preprint,12pt]{elsarticle}
\usepackage{hyperref}

\usepackage{multirow}
\usepackage{appendix}
\usepackage{graphicx}
\usepackage{subfigure}
\usepackage{enumitem}
\usepackage{color}
\usepackage{url}
\usepackage{bm}
\usepackage{amsmath}
\usepackage{amsfonts}
\usepackage{array}
\usepackage{framed} 
\usepackage[framed]{ntheorem}
\newframedtheorem{frm-thm}{Theorem}
\newcommand{\PreserveBackslash}[1]{\let\temp=\\#1\let\\=\temp}
\newcolumntype{C}[1]{>{\PreserveBackslash\centering}p{#1}}
\newcolumntype{R}[1]{>{\PreserveBackslash\raggedleft}p{#1}}
\newcolumntype{L}[1]{>{\PreserveBackslash\raggedright}p{#1}}
\newcommand{\tabincell}[2]{\begin{tabular}{@{}#1@{}}#2\end{tabular}}
\begin{document}

\begin{frontmatter}

\title{Resource Mention Extraction for MOOC Discussion Forums}
\tnotetext[mytitlenote]{}


\author[mymainaffiliation,mysecondaffiliation]{Ya-Hui An}

\author[mysecondaffiliation,myfourthaffiliation]{Liangming Pan \corref{mycorrespondingauthor}}
\cortext[mycorrespondingauthor]{Corresponding author}
\ead{peterpan10211020@gmail.com}

\author[mysecondaffiliation,mythirdaffiliation]{Min-Yen Kan}
\author[mymainaffiliation]{Qiang Dong}
\author[mymainaffiliation]{Yan Fu}

\address[mymainaffiliation]{School of Computer Science and Engineering, University of Electronic Science and Technology of China, Chengdu, Sichuan, P.R. China}
\address[mysecondaffiliation]{Web IR / NLP Group (WING), National University of Singapore, Singapore}
\address[myfourthaffiliation]{NUS Graduate School for Integrative Sciences and Engineering, National University of Singapore, Singapore}
\address[mythirdaffiliation]{NUS Institute for Application of Learning Science and Educational Technology (ALSET), Singapore}

\begin{abstract}

In discussions hosted on discussion forums for Massive Online Open Courses (MOOCs), references to online learning resources are often of central importance.  They contextualize the discussion, anchoring the discussion participants' presentation of the issues and their understanding.  However they are usually mentioned in free text, without appropriate hyperlinking to their associated resource.  
Automated learning resource mention hyperlinking and categorization will facilitate  discussion and searching within MOOC forums, and also benefit the contextualization of such resources across disparate views. 
We propose the novel problem of learning resource mention identification in MOOC forums; i.e., to identify resource mentions in discussions, and classify them into pre-defined resource types. 

As this is a novel task with no publicly available data, we first contribute a large-scale labeled dataset -- dubbed the Forum Resource Mention (FoRM) dataset -- to facilitate our current research and future research on this task. FoRM contains over $10,000$ real-world forum threads in collaboration with Coursera, with more than $23,000$ manually labeled resource mentions. 

We then formulate this task as a sequence tagging problem and investigate solution architectures to address the problem.  Importantly, we identify two major challenges that hinder the application of sequence tagging models to the task: (1) the diversity of resource mention expression, and (2) long-range contextual dependencies. We address these challenges by incorporating character-level and thread context information into a LSTM--CRF model. First, we incorporate a \textit{character encoder} to address the out-of-vocabulary problem caused by the diversity of mention expressions. Second, to address the context dependency challenge, we encode thread contexts using an RNN-based \textit{context encoder}, and apply the attention mechanism to selectively leverage useful context information during sequence tagging.  Experiments on FoRM show that the proposed method improves the baseline deep sequence tagging models notably, significantly bettering performance on instances that exemplify the two challenges. 
\end{abstract}


\end{frontmatter}

\section{Introduction}
\label{sec:intro}
\noindent

\begin{figure}[ht]
	\centering
	\includegraphics[width=\textwidth]{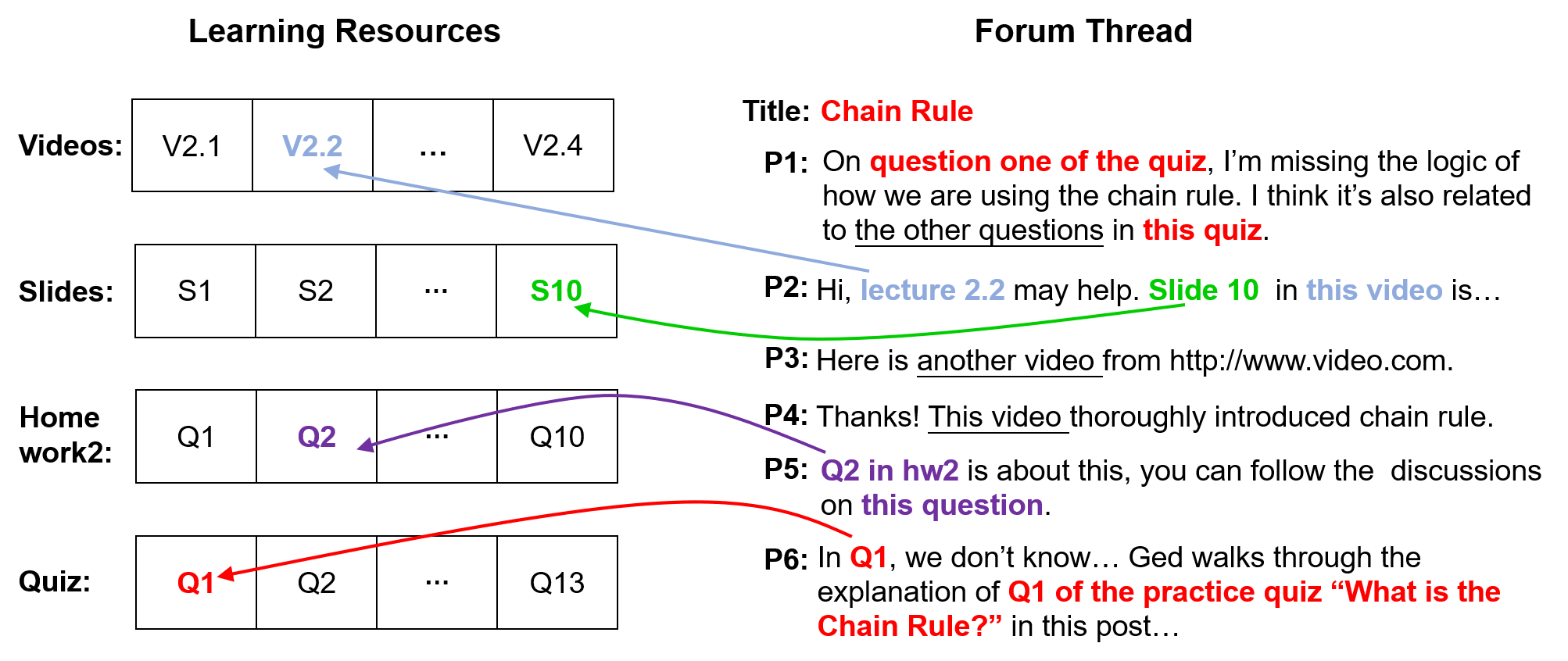}
\caption{An example of resource mention identification. The left shows the learning resources of a course, the right is a forum thread. $P1-P6$ are six posts in the thread. Resource mentions are marked in bold, with same color refer to the same learning resource. The underlined text are not valid mentions.}
	\label{fig:context-reference-problem}
\end{figure}

With the efforts towards building an interactive online learning environment, discussion forum has become an indispensable part in the current generation of MOOCs. In discussion forums, students or instructors could post problems or instructions directly by starting a thread or posting in an existing thread. During discussions, it is natural for students or instructors to refer to a learning resource, such as a certain quiz, this week's lecture video, or a particular page of slides. These references to resources are called \emph{resource mentions}, which compose the most informative parts among a long thread of posts and replies. 
The right side of Figure~\ref{fig:context-reference-problem} shows a real-world forum thread from Coursera\footnote{Coursera (\url{https://www.coursera.org/}) is one of the largest MOOC platforms in the world. }, in which resource mentions are highlighted in bold, with same color refer to the same resource on the left. From this example, we find that if we identify and highlight resource mentions in forum threads, it will greatly facilitate learners to efficiently seek for useful information in discussion forums, and also establish a strong linkage between a course and its forum.

We propose and study the problem of \emph{resource mention identification} in MOOC forums. Specifically, given a thread from MOOC discussion forum, our goal is to automatically identify all resource mentions present in this thread, and categorize each of them to its corresponding resource type. For resource types, we adopt the categorization proposed in ~\cite{an2018muir}, where learning resources are categorized into videos, slides, assessments, exams, transcripts, readings, and additional resources. 

Our task can be formulated as a sequence tagging problem. Given a forum thread as a word sequence $\mathcal{T} = \{ w_1, \cdots, w_n \}$, we apply a sequence tagging model to assign a tag $t_i$ to each word $w_i$, where $t_i$ represents either the  Beginning, Inside or Outside (BIO) of a certain type of resource mention (e.g., the tag ``Videos\_B" for $w_i$ indicates that $w_i$ is the first word of a resource mention with type ``Videos"). To train a sequence tagger, we need a large amount of labeled resource mentions in MOOC forums. However, to the best of our knowledge, no public labeled dataset is available since we are the first to investigate this task. To closely investigate this problem and also facilitate the following research on this task, we manually construct a large-scale dataset, namely \textit{Forum Resource Mention (FoRM)} dataset, in which each example is a forum post with labeled resource mentions. We first crawl real-world forum posts from Coursera, and then perform human annotations to identify resource mentions and their resource types. During the annotation, we find that resource mentions are hard to be identified even for human annotators. Compared with some well-studied sequence tagging problems such as POS tagging~\cite{brill1992simple,brants2000tnt}, and Named Entity Recognition (NER)~\cite{nadeau2007survey,lample2016neural,chiu2016named}, resource mention identification in MOOC forums poses several unique challenges. 
 
The most challenging issue is the \textit{context dependency}. Compared with other sequence tagging tasks such as POS tagging and NER, in which lexical patterns or local contexts serves as strong clues for identification, resource mention identification usually requires an understanding of the whole context in the thread. For example, in Figure~\ref{fig:context-reference-problem}, both the post \emph{P2} and \emph{P4} contain the mention ``this video". The mention in \emph{P2} is a valid resource mention, as it refers to a specific resource (\emph{Video 2.2}) within the course. However, in \emph{P4}, ``this video" actually refers to an external resource, thus is not a valid resource mention. As another example, the mention ``the other questions" in \emph{P1} is also an invalid resource mention, because it makes a general reference to the quiz questions. These examples reflect some of the typical scenarios in MOOC forums, in which the identification deals with long-range context dependencies, and require an in-depth understanding of the thread context. 
Another challenge comes from the \textit{variety of expressions}. Since the discussion forum is a colloquial communication environment, it is often filled with typos, abbreviations, compound words, new words, and other words that are not included in the dictionary, i.e., Out-of-Vocabulary (OOV) words. As shown in the post \emph{P6} of Figure~\ref{fig:context-reference-problem}, the word ``Q1" is a valid resource mention but also an OOV word. Identifying ``Q1" requires not only the context, but also an understanding of character-level semantics (e.g., ``Q" stands for ``Question"), which further increases the difficulty of this task.

We propose to add a \textit{character encoder} and a \textit{context encoder} to LSTM--CRF~\cite{huang2015bidirectional}, a state-of-the-art model for sequence tagging, to address the above challenges. 
First, to better capture the semantics of OOV words caused by the variety of expressions, we incorporate \textit{Character Encoder} to the original LSTM--CRF model, which encodes character-level information via LSTMs. This helps us better capture the correlation between abbreviations (e.g., ``Q1" and ``Q2") and the prefix or postfix information (e.g., ``dishdetail.html"). 
As for the context dependency problem, we need an effective way to leverage thread contexts, since LSTM--CRF usually has a hard time dealing with long-range context dependencies. To resolve this problem, we propose to add an attentive-based \textit{Context Encoder,} which encodes each context sentence with LSTMs, and selectively attends to useful contexts using the attention mechanism~\cite{bahdanau2014neural} during the decoding process of sequence tagging. 

Based on the constructed FoRM dataset, we subsequently evaluate the performance of different sequence tagging models, and conduct further analysis on how the proposed method solves the major challenges in resource mention identification. We evaluate the models on two versions of FoRM datasets: a medium-scale version (FoRM-M), which contains around 9,000 annotated resource mentions, and has high agreement between human annotators; a large-scale version (FoRM-L), which contains more than 25,000 annotated resource mentions, but with relatively lower annotation agreement. The resource mentions in FoRM-M are easier to identify from surface forms (e.g., ``Week 2 Quiz 1"); while mentions in FoRM-L are more ambiguous and dependent on the context. The experimental results show that our incremental LSTM--CRF model outperforms the baselines on both FoRM-M and FoRM-L, with noticeable effects on alleviating the above two challenges via incorporating character encoder and context encoder.  

The main contributions of this paper can be summarized as follows:
\begin{itemize}
\item The first attempt, to the best of our knowledge, to systematically investigate the problem of resource mention extraction in MOOC forums.
\item We propose an incremental model of LSTM--CRF that incorporates character encoder and context encoder, to solve the expression variety and context dependency problems. The model achieves an average improvement $F_1$ score of 3.16\% ({\it c.f.} Section~\ref{sec:ExperimentResults}) over LSTM--CRF. 

\item We construct a novel large-scale dataset, FoRM, from forums in Coursera, to evaluate our proposed method. 
\end{itemize}

The rest of the paper is organized as follows: In Section~\ref{sec:relatedwork}, we will first discuss some related works. In Section~\ref{sec:dataset}, we will introduce our dataset, FoRM. In Section~\ref{sec:method}, we formalize the problem, and illustrate our proposed model. We will provide the experimental results and analysis of the proposed method in Section~\ref{sec:experiment}. Finally, Section~\ref{sec:conclusion} will summarize the paper and discuss future research directions. 

\section{Related Works} 
\label{sec:relatedwork}

The task of resource mention identification can be regarded as a twin problem of \textit{named entity recognition} and \textit{anaphora resolution}, and we will elaborate both in the following.  

\subsection{Named Entity Recognition}

Despite some works have investigated extracting key concepts in MOOCs~\cite{wang2015constructing,pan2017course,pan2017prerequisite}, our work is different because the objective of our task is to jointly identify the position and type of resource mentions from plain texts. Therefore, it is more similar to \textit{Named Entity Recognition (NER)}, which seeks to locate named entities in texts and classify them into pre-defined categories. Neural sequence tagging models have become the dominate methodology for NER since the emerge and flourish of deep learning. Hammerton~\cite{hammerton2003named} attempted a single-direction LSTM network to perform sequence tagging, and Collobert et al.~\cite{collobert2011natural} employed a deep feed-forward neural network for NER, and achieved near state-of-the-art results. However, these NER models only utilize the input sequence when predicting the tag for a certain time-step, but ignoring the interaction between adjacent predictions. 
To address this problem, Huang et al.~\cite{huang2015bidirectional} proposed to add a CRF layer on top of a vanilla LSTM sequence tagger. This LSTM--CRF model has achieved the state-of-the-art results for NER when using the bidirectional LSTM (BLSTM). 

One problem of LSTM--CRF is that it only captures the word-level semantics. This causes a problem when intra-word morphological and character-level information are also very important for recognizing named entities. Recently, Santos et al.~\cite{dos2015boosting} augmented the work of Collobert et al.~\cite{collobert2011natural} with character-level CNNs. Chiu and Nichols ~\cite{chiu2016named} incorporated the character-level CNN to BLSTM and achieved a better performance in NER. In our task, resource mention identification, the widely existing OOV words, such as ``Q1'', ``Q2'', ``hw2'' in Figure~\ref{fig:context-reference-problem}, greatly increase the difficulty of capturing word-semantics. Therefore, we also incorporate the character-level semantics by proposing a character encoder via LSTM. 

However, incorporating character embeddings is insufficient for resource mention identification, as this task is different from NER with respect to the reliance on long-range contexts. Compared to NER, which typically requires limited context information, resource mention identification is a more context-dependent task. A common scenario is to judge whether a pronoun phrase, such as ``this video'', refers to a resource mention or not. For example, to understand that ``this video" in \textit{P4} of Figure~\ref{fig:context-reference-problem} actually does not refer to any resource within the course requires the contexts from at least \textit{P2}, \textit{P3} and \textit{P4}. In this case, this problem is more related to \textit{Anaphora Resolution}, which is another challenging problem in NLP.  

\subsection{Anaphora Resolution}

In computational linguistics, anaphora is typically defined as references to items mentioned earlier in the discourse or ``pointing back'' reference as described by ~\cite{mitkov1999multilingual}. \textit{Anaphora Resolution (AR)} is then defined as resolving anaphora to its corresponding entities in a discourse. Resolving repeated references to an entity is similar to differentiating whether a mention is a valid resource mention within the course. 

Most of the early AR algorithms were dependent on a set of hand-crafted rules. These early methods were a combination of salience, syntactic, semantic and
discourse constraints to do the antecedent selection. In 1978, Hobbs et al.~\cite{hobbs1978resolving} firstly combined
the rule-based, left to right breadth-first traversal of the syntactic parse tree of a sentence with selectional constraints to search for a single antecedent. Lappin et al.~\cite{lappin1994algorithm} discussed a discourse model to solve the pronominal AR. Then the centering theory~\cite{grosz1995centering,walker1998centering} was proposed as a novel algorithm used to explain phenomenon like anaphora using discourse structure. During the late nineties, the research in AR started to shift towards statistical and machine learning algorithms~\cite{aone1995evaluating,lee2017scaffolding,mitkov2002new,ge1998statistical}, which combines the rules or constraints of early works as features.  
Recently, the relevant research shifted to deep learning models for \textit{Coreference Resolution (CR)}, which includes AR as a sub-task. Wiseman et al.~\cite{wiseman2015learning} designed mention ranking model by learning different feature representations for anaphoricity detection and antecedent ranking by pre-training on these two individual subtasks~\cite{sukthanker2018anaphora}. Later, they proved that coreference task can benefit from modeling global features about entity clusters~\cite{wiseman2016learning}. Meanwhile, Clark et al.~\cite{clark2016deep} proposed another cluster ranking model to derive global information. Up to now, the state-of-the-art model was proposed by ~\cite{lee2017end}, an end-to-end CR system that jointly modeled mention detection and CR.

Most of the AR works take as input the candidate key phrases extracted from the discourse, and then resolve these phrases to entities by casting the problem as either a classification or ranking task. However, our task is defined as a sequence tagging problem, which requires anaphora resolution implicitly when predicting the type of an ambiguous resource mention. In our model, we incorporate a context encoder to implement a mechanism of sequence-to-sequence tagging with attention to help the model to learn anaphora resolution within the contexts implicitly during training.  

\section{The FoRM Dataset}
\label{sec:dataset}

In this section, we introduce the construction of our experimental dataset, i.e., Forum Resource Mention  (FoRM) dataset. To the best of our knowledge, there is no publicly available dataset that contains labeled resource mentions in MOOC forums.
We construct our dataset via a three-stage process: (1) data collection, (2) data annotation, and (3) dataset construction. 

\subsection{Data Collection}
Our data comes from Coursera, one of the largest MOOC platforms in the world. Coursera was founded in 2012 and up to August 2018, it has offered more than 2,700 courses and attracted about 33 million registered learners. Each course has a discussion forum for students to post/reply questions and to communicate with each other. Each forum contains all the threads started by students or instructors, which consists of one \textit{thread title} (main idea of a problem), one or more \textit{thread posts} (details about the problem) and replies (see Figure~\ref{fig:context-reference-problem} as an example). 

As the distribution of resource mentions may vary for courses in different domains, we consider a wide variety of course domains when collecting the data. Specifically, we collect the forum threads from $142$ completed courses in $10$ different domains\footnote{The selected domains are: {\it `Arts and Humanities'}, {\it `Business'}, {\it `Computer Science'}, {\it `Data Science'}, {\it `Language Learning'}, {\it `Life Sciences'}, {\it `Math and Logic'}, {\it `Personal Development'}, {\it `Physical Science and Engineering'} and {\it Social Sciences}}. Note that in Coursera, each course may have multiple sessions; each session is an independent learning iteration of the course, with a fixed start date and end date (e.g., ``Machine Learning" (from 2018-08-20 to 2018-12-20)). Different sessions of a course may have different organization and notation systems for the same set of learning resources, which involves ambiguity if we consider them all. Therefore, we only select the latest completed session for each course, resulting a total number of $102,661$ posts\footnote{Our data was collected at January 31, 2017, and we are in partnership with Coursera at the time of the dataset collection.}. Finally, we exclude the posts that belong to the ``General Discussion'' and ``Meet \& Greet'' forums, which are unlikely to contain resource mentions, and only select the posts in ``Week Forums", as they are designed for ``Discuss and ask questions about Week X". This gives us a data collection of $84,945$ posts from $11,679$ different forum threads. 

\subsection{Data Annotation}
Based on the above collected data, we then manually annotate resource mentions for each thread. We employ $16$ graduate students from technical backgrounds to annotate the data. As mentioned before, our data collection consists of $11,679$ forum threads from $142$ courses; each thread is a time-ordered list of posts, including thread title and a series of thread/reply posts. We split the $11,679$ threads into $8$ portions, and assign each portion to $2$ annotators. For simplicity of annotation, for each thread, we concatenate all contents of its posts, to get a single document of sentences for annotation. For each thread document, the task of the annotator is to identify all the resource mentions in the document, and tag each of them with one of the pre-defined $7$ resource types defined in Section ~\ref{sec:intro} (refer to Table~\ref{table:tab-types-resource} for details). We define a resource mention as any one or more consecutive words in a sentence that represents an unambiguous learning resource in the course. We use the \textit{brat rapid annotation tool}\footnote{\url{http://brat.nlplab.org/}}, an online environment for collaborative text annotation, which is widely used in entity, relation and event annotations~\cite{ohta2012open,kim2009overview,kim2011overview}, as our annotation platform. 

\begin{table}
	\begin{center}
		\begin{tabular}{  c | c | c | c | c | c} \hline
		Resource Type 	&  Group 1  & Group 2 &  Intersection & Union & $P_{pos}$\\ \hline
        \hline
		Assessments & 8,047 & 8,520 & 5,451 & 11,116 &0.658\\ \hline
        Exams & 1,891 & 3,624 & 1,146 & 4,369 &0.416\\ \hline
        Videos & 1,852 & 3,037 & 1,236 & 3,653&0.506\\ \hline
        Coursewares & 3,281 & 4,286 & 1,557 & 6,010&0.412\\ \hline
        \hline
        Total & 15,071 &  19,467 & 9,390 & 25,148 & 0.544\\ \hline
		\end{tabular}
	\end{center}
\caption{Annotation result from Group 1 and Group 2 on Assessments, Exams, Videos and Coursewares. Coursewares = Readings, Slides, Transcripts, Additional Resources. $P_{pos}$ is the Positive Specific Agreement.}
\label{tab-annotation-result}    
\end{table} 

To help annotators better understand the above process and relevant concepts, we conduct an one-hour training for annotators; the complete training process is documented in~\ref{sec:appendix}. Then, we start the real annotation; the whole annotation process takes around one month. In the end, each thread is doubly annotated, and we denote the two copies of the annotated data as Group $1$ and Group $2$, respectively. Table ~\ref{tab-annotation-result} summarizes the the number of annotated resource mentions for each resource type. Note that we integrate the $4$ resource types representing teaching materials, i.e., {\it `Readings'}, {\it `Slides'}, {\it `Transcripts'}, and {\it `Additional Resources'}, into one single resource type {\it `Coursewares'}, to form a dataset with more balanced training examples for each class. 

\begin{table}
\small
\begin{center}
\begin{tabular}{  c | c | c }\hline
Type & Description & Notation \\ \hline
\hline
Agree & text span overlaps, and annotated type same & $AG$ \\ \hline
Type-Disagrees & text span overlaps, but annotated type different & $TD$\\ \hline
G1-Only & the annotation exists only in Group 1 & $G1$ \\ \hline
G2-Only & the annotation exists only in Group 2 & $G2$ \\ \hline
\end{tabular}
\end{center}
\caption{Four possible cases when comparing the annotation results of Group 1 and 2. }
\label{tab-annotation-comparation}   
\end{table}

To evaluate the inter-annotator agreement between two groups, we use the \textit{Positive Specific Agreement}~\cite{hripcsak2005agreement}, a widely-used measure of agreement when the positive cases are rare compared with the negative cases. 
In summary, there are $4$ possible cases when comparing the result of the annotated mentions between Group 1 and Group 2, summarized in Table~\ref{tab-annotation-comparation}. For example, $a$ denotes the number of cases that both groups agree are resource mentions and also have an agreement about its type. Based on all the conditions listed in Table~\ref{tab-annotation-comparation}, the calculation of the positive specific agreement (denoted as $P_{pos}$) between two groups' annotations is given in Equation~\ref{pos-a}. The agreement scores for different resource types are shown in the column $P_{pos}$ of Table~\ref{tab-annotation-result}.

\begin{equation}\footnotesize
\label{pos-a}
P_{pos} = \frac{2 \times AG}{2 \times AG + (TD + G1)+(TD + G2)}
\end{equation}

To give an explanation for $P_{pos}$ values to better understand whether our annotation achieves an acceptable agreement, we analyze the value of $P_{pos}$ by referring to Kappa coefficient, because~\cite{hripcsak2005agreement} proves that $\kappa$ approaches the positive specific agreement when the number of negative cases grows large, which is exactly our case. We find that the $P_{pos}$ value for \textit{Exams}, \textit{Videos} and \textit{Coursewares} are in the range of \textit{moderate agreement}\footnote{The values for $\kappa$: [-1, 0): less than chance agreement; 0: random; [0.01, 0.20]: slight agreement; [0.21, 0.40]: fair agreement; [0.41, 0.60]: moderate agreement; [0.61, 0.80]: substantial agreement; and [0.81, 0.99] almost perfect agreement; 1: perfect agreement.}, and for \textit{Assessments}, the value shows a \textit{substantial agreement}~\cite{viera2005understanding}. The possible reasons that the agreement for \textit{Assessments} is higher than the other types are: 1) samples for four types of resource are unbalanced; the ratio of Assessments is higher than others, thus has a lower annotation bias; 2) Assessments is easier for annotators to distinguish compared to other types of resource. In summary, the overall annotation result achieves a moderate agreement between two group of annotators. 

\subsection{Dataset Construction}

Based on the annotation results, we construct two versions of datasets with different characteristics. First, to provide a dataset with high-quality resource mentions, we only use the ``Agree'' cases in Table~\ref{tab-annotation-comparation} as the ground-truth resource mentions to construct the \textit{FoRM-M} dataset. For the ``Agree'' case, we joint the text spans of annotated mentions from Group 1 and Group 2 as the ground truth. For example, if the annotated mentions are ``the video 1'' (Group 1) and ``video 1 of week 2'' (Group 2), we create a ground-truth of ``the video 1 of week 2'' by unioning the texts. In this way, we tend to obtain more specific mentions (e.g., ``the video 1 of week 2'') rather than general ones (e.g., ``video 1"). The number of ``Agree'' resource mentions is $9,390$ as shown in the column ``Intersection'' in Table~\ref{tab-annotation-result}. We also construct a larger but relatively more noisy dataset, namely \textit{FoRM-L}, by using the ``Agree'', ``G1-Only'', and ``G2-Only'' cases as ground-truths, which represents a ``union" of the annotations from the two groups. The statistics are shown in the ``Union'' column of Table~\ref{tab-annotation-result}. 

\begin{table}\small
	\begin{center}
		\begin{tabular}{ l c|c|c } \hline 
         \multicolumn{2}{c|}{ Dataset } & FoRM-M& FoRM-L \\ \hline 
         \hline
         \multicolumn{2}{c|}{ \# Examples} & 8,390 & 19,952 \\ \hline
         \multicolumn{2}{c|}{ \# Tokens} & 150,597 & 395,958\\ \hline
         \multicolumn{2}{c|}{ \# Average Length} & 17.95 & 19.84\\ \hline
         \multirow {9}{*}{\# Tags}  & Coursewares\_B & 1,398 & 5,183 \\ \cline{2-4}
         & Coursewares\_I & 1,273 & 1,273\\ \cline{2-4}
         & Exams\_B & 1,094 & 3,989 \\ \cline{2-4}
         & Exams\_I & 1,901 & 4,166\\ \cline{2-4}
         & Assessments\_B & 5,202 & 10,432 \\ \cline{2-4}
         & Assessments\_I & 7,359 & 12,885\\ \cline{2-4}
         & Videos\_B & 1,223 & 3,403 \\ \cline{2-4}
         & Videos\_I& 2,121 & 4,670 \\ \cline{2-4}
         & O & 129,026 & 346,693\\ \hline        
        \end{tabular}
	\end{center}
\caption{Statistics of the Forum Resource Mention Dataset (FoRM).}
\label{tab-dataset}    
\end{table} 

As mentioned in Section~\ref{sec:intro}, we formulate the task of resource mention identification in MOOC forums as a sequence tagging problem. Therefore, we associate each word in the dataset with a corresponding tag, based on the ground-truth we obtained in the previous step. A word is associated with the \textit{Beginning (B)}/ \textit{Inside (I)} tag if it is the beginning/inside of a resource mention with type $T$, denoted as $T\_B/I$. Otherwise, the \textit{Outside (O)} tag is assigned to the word. 

The statistics of the constructed datasets are shown in Table~\ref{tab-dataset}, where {\it \# Examples} is the total number of sentences containing at least one resource mention, {\it \# Tokens} is the total number of words in the dataset. {\it \# Average Length} denotes the average number of words in a sentence. The total number of B-tags (e.g., Coursewares\_B) and I-tags (e.g., Exams\_I) for different resource types, as well as the number of O-tags, are also listed in the table. 
 
\section{Methods}
\label{sec:method}
We present our neural model for identifying and typing resource mentions in MOOC forums. We first formulate the problem and then present the general architecture of the proposed model. Followed by that, we introduce the major components of our model in detail in the remaining sections.

\subsection{Problem Formulation}
\label{sec:problem_formulation}

We first introduce some basic concepts, and formally define the task of resource mention identification in MOOC forums. \\

\noindent \textbf{\textit{Definition 1}} \textbf{(Post)} A post $P$ is the smallest unit of communication in MOOC forums that contains user-posted contents. Each post is composed of the text contents written by the user, and some associated meta-data such as user ID, posting time etc. In our task, we focus on extracting resource mentions from text contents; thus we simply formulate a post as a sequence of sentences, i.e., $P = \{ s_1, \cdots, s_{|P|} \}$, where each sentence is a word sequence $s = \{ w_1, \cdots, w_{|s|} \}$. \\

\noindent \textbf{\textit{Definition 2}} \textbf{(Thread)} Typically, a thread $T$ in MOOC forums is composed of a thread title $t$, an initiating post $I$, and a set of reply posts $R$~\cite{bhatia2010adopting}. Initiating post is the first post in the thread and initiates discussions. All other posts in a thread are the reply posts that participate in the discussion started by the initiating post. For simplicity, we do not differentiate between the initiating post and the reply posts, and we also treat the thread title as a special post $P_0$. In this case, a thread $T$ can be represented as an ordered list of posts, i.e., $T = \{ P_0, P_1, \cdots, P_{|T|} \}$.
A thread $T$ with $n$ posts can be unfolded as a long document of $N$ sentences $T = \{ s_1, \cdots, s_N : s_i \in P_{I(i)} \}$, where $I(i)$ is the index of the post that sentence $s_i$ belongs to. \\

\noindent \textbf{\textit{Definition 3}} \textbf{(Resource Mention)} A \textit{course} $C$ in MOOCs is defined as a set of resources, where each \textit{resource} represents a specific learning resource/material in $C$ (e.g., ``Video 2.1"), and is associated with a \textit{resource type} (e.g., ``Video"). In a thread that belongs to course $C$, we define any semantically complete single/multi-word phrase that represents a resource of $C$ as a \textit{resource mention} (e.g., ``the first video of chapter 2"). \\

\noindent \textbf{\textit{Definition 4}} \textbf{(Resource Mention Identification)} The task of resource mention identification in MOOC threads is defined as follows: Given a thread $T$ in the discussion forums of course $C$, the objective is to identify all resource mentions appearing in $T$, and for each identified resource mention, to categorize it into one of the pre-defined resource types. \\

This task involves identifying both the location and the type of a resource mention, so it can be formulated as a sequence tagging problem.
Specifically, given a thread $T$, our task is to assign a tag $t$ to each word $w \in T$. The tag $t$ can be either $\mathcal{T}_B$ (the begining of a resource mention of type $\mathcal{T}$), $\mathcal{T}_I$ (inside a resource mention of type $\mathcal{T}$), or $\mathcal{O}$ (outside any resource mention).
Under this problem formulation, state-of-the-art sequence tagging models, such as LSTM--CRF, can be applied to our task. However, they suffer from the two major challenges discussed in Section~\ref{sec:intro}. Therefore, we propose an incremental neural model based on LSTM--CRF to address the challenges. In the following sections, we will introduce our model in detail, and more specifically, discuss how we address the above two challenges by incorporating the context encoder and the character encoder.

\subsection{General Architecture}
\label{sec:framework}
A thread $T$ with $n$ posts is unfolded as a sequence of $N$ sentences $T = \{ s_1, \cdots, s_N \}$, where $s_i$ is the $i$-th sentence in the entire thread $T$. 
Given $T$ as input, our model performs sentence-level sequence tagging for each sentence in the thread $T$. Specifically, to decode the sentence $s_i \in T$, we consider all or part of the previous sentences of $s_i$ as its \emph{contexts}, denoted as $C_i$. Then, our goal is to learn a model that assigns each word in $s_i$ with a tag; we denote the output tag sequence as $t_i$. Therefore, our model essentially approximates the following conditional probability.
\begin{equation}
\label{equ:framework}
p(Y\: | \: T \: ; \: \Theta) = \prod_{i=1}^{N} p(t_i \: | \: s_i, C_i \: ; \: \Theta)
\end{equation}
where $\Theta$ is the model parameters, and $p(t_i \: | \: s_i, C_i \: ; \: \Theta)$ denotes the conditional probability of the output tag sequence $t_i$ given the sentence $s_i$ and its context $C_i$.

\begin{figure}[ht]
	\centering
	\includegraphics[width=\textwidth]{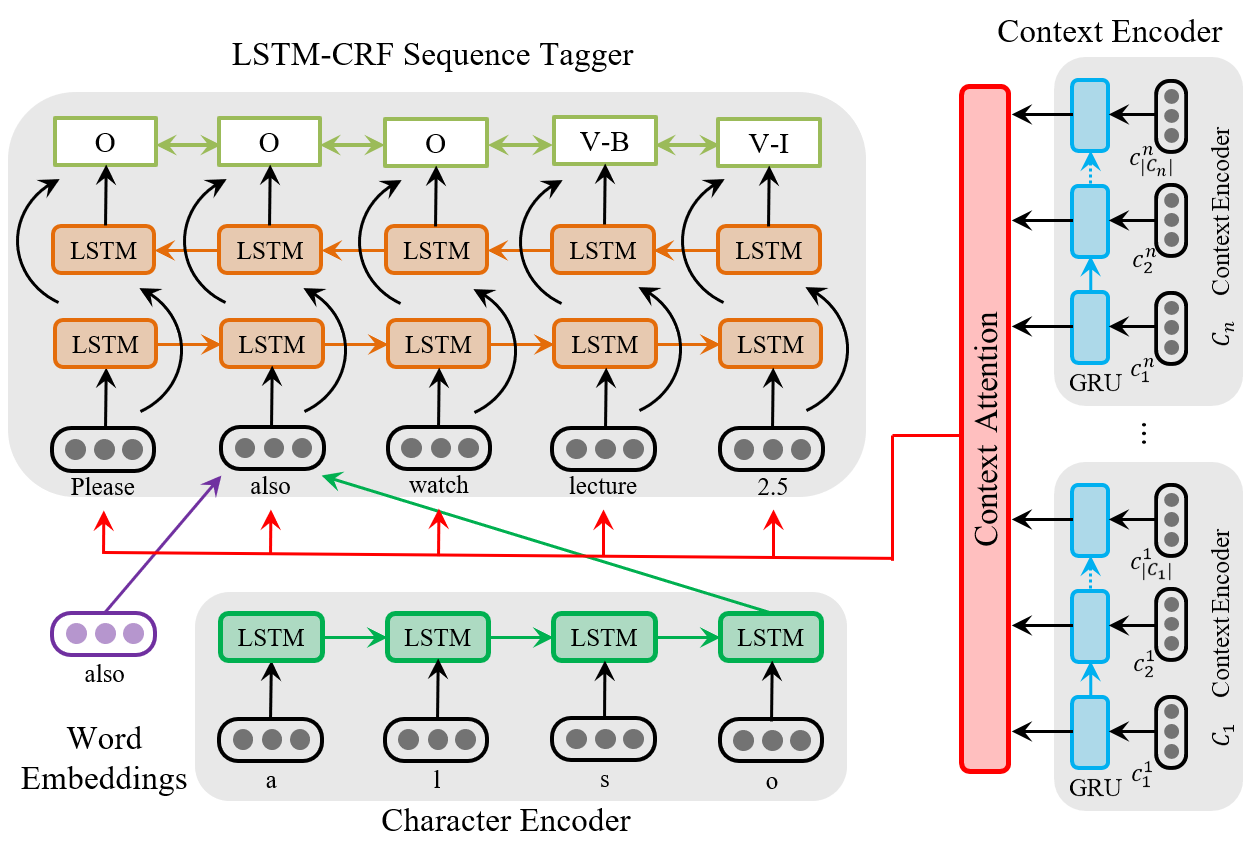}
	\caption{The general architecture of the proposed model. Our model consists of three parts: Context Encoder, Character Encoder, and LSTM--CRF, which are shaded in gray. }

\label{fig:general-architecture}
\end{figure}

To model the conditional probability $p(t_i \: | \: s_i, C_i \: ; \: \Theta)$, our model includes three components: (1) the context encoder, (2) the character encoder, and (3) the attentive LSTM--CRF tagger. Figure~\ref{fig:general-architecture} shows the framework of our proposed neural model. First, to encode the context information $C_i$, we incorporate the \emph{context encoder}: a set of recurrent neural network (RNN) to encode each context sentence (Section~\ref{sec:ContextEncoder}). Our context encoder is generic to any textual contexts that can be additionally provided (e.g., from external resources), while in our model, we use the previous sentences of the thread as the context, to address the context dependency problem proposed in Section~\ref{sec:intro}. To alleviate the OOV challenge in our task, we employ the \emph{character encoder} to build word embeddings using BLSTMs~\cite{schuster1997bidirectional} over the characters (Section~\ref{sec:CharacterEncoder}). The character-level word embeddings are then combined with the word-level embeddings as inputs to our model. Finally, we use the BLSTM--CRF~\cite{huang2015bidirectional} to generate the output tag sequence. Different from the original model in~\cite{huang2015bidirectional}, we add an attention module~\cite{bahdanau2014neural} that acts over the encoded textual contexts (\emph{attentive LSTM--CRF tagger}), to make use of important context information during sequence tagging (Section~\ref{sec:LSTMTagger}).

\subsection{Context Encoder}
\label{sec:ContextEncoder}

As discussed in Section~\ref{sec:intro}, context information is crucial for identifying resource mentions. For the $i$-th sentence $s_i$ in the input thread $T$, a straightforward way is to use the \textbf{thread context}, which is to encode all the previous sentences of $s_i$ in $T$ as its context, i.e., $C_i = \{ s_1, \cdots, s_{i-1} \}$. The thread context contains complete information for inferring resource mentions in $s_i$, but also makes it harder for the model to learn the inherent patterns from these long and noisy contexts. We address this problem by introducing the attention mechanism into the decoding process, which will be further illustrated in Section~\ref{sec:LSTMTagger}. 

We denote the thread context $C$ as a sequence of $m$ sentences $C = \{ c_1, \cdots, c_m \: | \: c_i = (c^i_1, \cdots, c^i_{|c_i|}) \}$, where $c_j^i$ represents the one-hot encoding of the $j$-th token in the $i$-th context sentence $c_i$, and $|c_i|$ is the length of the sentence $c_i$ ({\it cf} Figure~\ref{fig:general-architecture}, each gray block represents the encoding of a sentence $c_i$ in context $C$). We employ the method in~\cite{elsahar2018zero} to use a set of $m$ Gated Recurrent Neural Networks (GRU)~\cite{cho2014learning} to encode each of the context sentence separately:
\begin{equation}
\label{equ:context_encoder}
h^{c_i}_j = \text{GRU}_i \left( \mathbf{E_c} \: c_j^i \: , \: h_{j-1}^{c_i} \right)
\end{equation}
where $\text{GRU}_i$ denotes the GRU used to encode the $i$-th context sentence $c_i$, $\mathbf{E_c}$ is the input word embedding matrix, and $h^{c_i}_j \in \mathbb{R}^{H_c}$ is the GRU hidden state in the $j$-th time step, which is determined by the input token $c_j^i$ and the previous hidden state $h_{j-1}^{c_i}$. We concatenate the last hidden state $h_{|c_i|}^{c_i}$ for each encoded context sentence $c_i$ to obtain our \textbf{context vector} $h_c$ as follows:
\begin{equation}\label{equ:context_vector}
h_c = \left[ h_{|c_1|}^{c_1}; \cdots ; h_{|c_i|}^{c_i} ; \cdots ; h_{|c_m|}^{c_m} \right]
\end{equation}

The context vector will further be used by the attention mechanism in Section~\ref{sec:LSTMTagger} to provide contextual information in the sequence tagging process.

\subsection{Character Encoder}
\label{sec:CharacterEncoder}

As discussed in Section~\ref{sec:intro}, our task suffers from the OOV problem, i.e., a large portion of words in forums (e.g., ``Q4") are not in the vocabulary.
This problem can be alleviated by incorporating the character-level semantics (e.g., the postfix ``.pdf" in the word ``intro.pdf"). In fact, introducing the character-level inputs to build word embeddings has already been proved to be effective in various NLP tasks, such as part-of-speech tagging~\cite{kim2016character} and language modeling~\cite{ling2015finding}. In our model, we build up a character encoder to encode character-level embeddings to fight against the OOV problem. For each word, we use bidirectional LSTMs to process the sequence of its characters from both sides and their final state vectors are concatenated. The resulting representation is then concatenated with the word-level embeddings to feed to the sequence tagger in Section~\ref{sec:LSTMTagger}.

We denote $\mathcal{V}_C$ as the alphabet of characters, including uppercase and lowercase letters as well as numbers and punctuation, with dimensionality in the low hundreds. The input word $w$ is decomposed into a sequence of characters $x_1, \cdots, x_{|w|}$, with each $x_i$ represented as an one-hot vector over $\mathcal{V}_C$. We denote $\mathbf{E_c} \in \mathbb{R}^{d_c \times \mathcal{V}_C}$ as the input character embedding matrix, where $d_c$ is the dimension of character embeddings. Given $x_1, \cdots, x_{|w|}$, a bidirectional LSTM computes the forward state $h^f_i$ by applying $h^f_i = LSTM(\mathbf{E_c} c_i, h^f_{i-1})$, and computes the backward state $h^b_i$ by applying $h^b_i = LSTM(\mathbf{E_c} c_{|w|-i+1}, h^b_{i-1})$. Finally, the input vector $v_w$ to the sequence tagger is the concatenation of word and character embeddings, i.e., $v_w = [ \mathbf{E_w} w\: ; \: h^f_{|w|} \: ; \: h^b_{|w|}]$.

\subsection{LSTM--CRF Tagger}
\label{sec:LSTMTagger}

After defining the input vector $v_w$ and the context vector $h_c$, we build up the attentive LSTM--CRF tagger to assign a tag to each word. Given a sentence with $n$ words $s = \{ w_1, \cdots, w_T \}$ in the input thread $T$ with context $C$, to obtain its tag sequence $l = \{ l_1, \cdots, l_T \}$, we are actually approximating the conditional probability $p(l_1, \cdots, l_T | w_1, \cdots, w_T, C)$. This can be effectively modeled by the LSTM--CRF tagger~\cite{huang2015bidirectional} in the following way.
\begin{equation}
\label{equ:LSTMCRF1}
p(l_1, \cdots, l_T | w_1, \cdots, w_T, C) = \frac{\exp(r(s, l | C))}{\sum_{l'}\exp(r(s,l' | C))}
\end{equation}
where $r(s, l | C)$ is a scoring function indicating how well the tag sequence $l$ fits the given input sentence $s$, given the context $C$. In LSTM--CRF, $r(s, l | C)$ is parameterized by a transition matrix $A$ and a non-linear neural network $f$, as follows:
\begin{equation}
\label{equ:LSTMCRF2}
r(s, l | C) = \sum_{t=1}^T \left ( A_{l_{t-1}, l_t} + f(w_t, l_t | C) \right )
\end{equation}
where $f(w_t, l_t | C)$ is the score output by the LSTM network for the $t$-th word $w_t$ and the $t$-th tag $l_t$, conditioned on the context $C$. The matrix $A$ is the \emph{transition score matrix}, $[A]_{ij}$ is the transition score from $i$-th tag to $j$-th for a consecutive time steps.

To model the score $f(w_t, l_t | C)$, we build a bidirectional-LSTM network with attention over the contexts $C$. In time step $t$, the current hidden state $h_t$ is updated as follows:
\begin{equation}
\label{equ:LSTMCRF3}
h_t = LSTM([v_{w_t}\: ; \: a^t_c], h_{t-1})
\end{equation}
where $v_{w_t}$ is input vector for word $w_t$, $a^t_c$ is the attended context vector of $h_c$ at time step $t$, which will be discussed in detail later. Then, the score $f(w_t, l_t | C)$ is computed through a linear output layer with softmax, as follows:
\begin{align}
\label{equ:LSTMCRF4}
o_t & = W_o h_t \\
f(w_t, l_t| C) & = \frac{\exp(o_{l_t, t})}{\sum_j \exp(o_{j,t})}
\end{align}
where $W_o$ is the matrix that maps hidden states $h_t$ to output states $o_t$.

\subsection{Context Attention on the Tagger}
\label{sec:AttentionMachanisim}

To effective select useful information from the contexts, we introduce an attention mechanism over all the hidden states of the context sentences $h_{|c_1|}^{c_1}, \cdots , h_{|c_i|}^{c_i} , \cdots , h_{|c_m|}^{c_m}$. We denote $\alpha^t_{i}$ as the scalar value determining the attention weight of the context vector $h_{|c_i|}^{c_i}$ at time step $t$. Then, the input context vector to the LSTM--CRF tagger $a^t_c$ is calculated as follows:
\begin{equation}
\label{equ:LSTMCRF5}
a^t_c = \sum_{i = 1}^m \alpha^t_{i} h_{|c_i|}^{c_i}
\end{equation}

Given the previous state of the LSTM $h_{t-1}$, the attention mechanism calculates the context attention weights $\alpha^t = \alpha^t_{1}, \cdots, \alpha^t_{m}$ as a vector of scalar weights, where $\alpha^t_{i}$ is calculated as follows:
\begin{align}
\label{equ:LSTMCRF6}
e^t_{i} & = v_a^{\top} \tanh (W_a h_{t-1} + U_a h_{|c_i|}^{c_i}) \\
\alpha^t_{i} & = \frac{\exp(e^t_i)}{\sum_j \exp(e^t_j)}
\end{align}
where $v_a, W_a, U_a$ are trainable weight matrices of the attention modules. Note that we actually calculate an attention over all context sentences, but not on the word level, which greatly reduce the scale of parameters. Another reason to use sentence-level attention is based on the observation that the useful information tends to appear coherently in one context sentence, rather than separated in different sentences.

\section{Experiments}
\label{sec:experiment}
\subsection{Baselines}

Since we formulate our task as a sequence tagging problem, to evaluate the performance of the proposed method, we conduct experiments on several widely-used sequence tagging models as follows:

\begin{itemize}
	\item \textbf{BLSTM}: the bidirectional LSTM network (BLSTM)~\cite{graves2013speech} has been widely used for sequence tagging task. In predicting the tag of a specific time frame, it can efficiently make use of past features (via forward states) and future features (via backward states). We train the BLSTM using back-propagation through time (BPTT)~\cite{boden2002guide} with each sentence-tag pair $(s, l)$ as a training example.
	\item \textbf{CRF}: Conditional Random Fields (CRF)~\cite{lafferty2001conditional} is a sequence tagging model that utilizes neighboring tags and sentence-level features in predicting current tags. In our implementation of CRF, we use the following features: (1) current word, (2) the first/last two/three characters of the current word, (3) whether the word is digit/title/in upper case, (4) the POS tag, (5) the first two symbols of the POS tag, and (6) the features (1)-(5) for the previous and next two words.
	\item \textbf{BLSTM--CRF}: As we illustrated in Section ~\ref{sec:LSTMTagger}, BLSTM--CRF ~\cite{huang2015bidirectional} is a state-of-the-art sequence tagging model that combines a BLSTM network with a CRF layer. It can efficiently use past input features via a LSTM layer and sentence level tag information via a CRF layer.
	\item \textbf{BLSTM--CRF--CE}: This model adds a character encoder (CE), as described in Section~\ref{sec:CharacterEncoder}, into the BLSTM--CRF model. It can be regarded as a simplified version of the proposed model, i.e., without the context encoder.
	\item \textbf{BLSTM--CRF--CE--CA}: The full version of the proposed method, i.e., an incremental model of BLSTM--CRF that takes into account the character-level inputs and the thread context information.
\end{itemize}

\subsection{Experimental Settings}

\noindent \textbf{Datasets.} We test LSTM, CRF, LSTM\textendash CRF, LSTM--CRF--CE and our model on both the FoRM-M and the FoRM-L datasets. For each dataset, we randomly split the data into 2 parts: $90\%$ for training and $10\%$ for testing. This results in $6,796$ training and $839$ testing examples for FoRM-M, and $16,160$ training and $1,996$ testing examples for FoRM-L. \\

\noindent \textbf{Setup.} For deep learning models, we set the size of the word representation to $200$, and initialize the word embedding matrix with pre-trained GloVe~\cite{pennington2014glove} vectors. In the LSTM--CRF--CE and our model, we set the dimensionality of characters to $64$. Each hidden state used in the LSTM and GRU is set to $256$. We train all models by stochastic gradient descent, with a minibatch size of $16$, using the ADAM optimizer. For the CRF model, we implement it using the \emph{keras-contrib}\footnote{https://github.com/keras-team/keras-contrib} package. To evaluate the overall performance, we use the micro-precison/recall/f1 score on all the resource mention tags, i.e., all tags excluding the $O$ tag, calculated as follows:
\begin{align}
\label{equ:microPRF}
\text{micro-}P & = \frac{\sum_{t \in L/O} TP_t}{\sum_{t \in L/O} (TP_t + FP_t)} \\
\text{micro-}R & = \frac{\sum_{t \in L/O} TP_t}{\sum_{t \in L/O} (TP_t + FN_t)} \\
\text{micro-}F_1 & = \frac {2 \times \text{micro-}R \times \text{micro-}P} {\text{micro-}R + \text{micro-}P}
\end{align}
\noindent where $L$ is the tag set, $TP_t$, $FP_t$ and $FN_t$ represents the number of true positive, false positive, and false negative examples for the tag $t \in L$, respectively.

\begin{table}\footnotesize
	\begin{center}
		\begin{tabular}{ c|c|c|c|c|c|c } \hline
			\multirow {2}{*}{Models}  & \multicolumn{3}{c|}{FoRM-M} & \multicolumn{3}{c}{FoRM-L} \\ \cline{2-7}
		    & Precision & Recall & $F_1$ Score & Precision & Recall & $F_1$ Score \\\hline \hline
           \multirow{2}{*}{BLSTM} & \multirow{2}{*}{71.53} & \multirow{2}{*}{47.32} & \multirow{2}{*}{56.96} & \multirow{2}{*}{64.46} & \multirow{2}{*}{48.04} & \multirow{2}{*}{55.05} \\
           &&&&&& \\ \hline
           \multirow{2}{*}{CRF} & \multirow{2}{*}{\textbf{78.08}} & \multirow{2}{*}{69.09} & \multirow{2}{*}{73.31} & \multirow{2}{*}{73.82} & \multirow{2}{*}{62.03} & \multirow{2}{*}{67.41} \\
           &&&&&& \\ \hline
		   \multirow{2}{*}{BLSTM--CRF} & \multirow{2}{*}{75.40} & \multirow{2}{*}{72.38} & \multirow{2}{*}{73.90} & \multirow{2}{*}{74.39} &
           \multirow{2}{*}{66.01}
           & \multirow{2}{*}{69.94} \\
           &&&&&& \\ \hline
		   \multirow{2}{*}{BLSTM--CRF--CE} & \multirow{2}{*}{73.76} & \multirow{2}{*}{77.62} & \multirow{2}{*}{75.64} & \multirow{2}{*}{\textbf{76.20}} &
           \multirow{2}{*}{70.21}
           & \multirow{2}{*}{73.08} \\
           &&&&&& \\ \hline
		   BLSTM--CRF--CE &\multirow{2}{*}{72.91}&\multirow{2}{*}{{\bf 79.20}}&\multirow{2}{*}{{\bf 75.92}}&\multirow{2}{*}{74.32}&\multirow{2}{*}{{\bf
           74.17}}&\multirow{2}{*}{{\bf 74.24}}\\
		   --CA &&&&&&\\\hline
		\end{tabular}
	\end{center}
	\caption{Overall performance of different methods on the FoRM dataset ($\%$). The best performances for each metric are in bold. }
	\label{tbl:final-result}
\end{table}

\subsection{Experimental Results}
\label{sec:ExperimentResults}

We train models using training data and monitor performance on validation data. During training, $10\%$ of training data are held out for validation ($10$-fold cross validation). The model is re-trained on the entire training data with the best parameter settings, and finally evaluated on the test data. For deep learning models, we use a learning rate of 0.01, and the training process requires less than 20 epochs to converge and it in general takes less than a few hours.

We report models' performance on test datasets in Table~\ref{tbl:final-result}, in which the best results are in bold cases. On both FoRM-M and FoRM-L dataset, BLSTM--CRF--CE--CA achieves the best $F_1$ score, which indicates the robustness and effectiveness of the proposed method. Specifically, we also have the following observations.

\begin{description}
  \item[(1)] BLSTM is the weakest baseline for both two data sets. It obtains relatively high precision but poor recall. When predicting current tags, BLSTM only considers the previous and post words, without making use of the neighboring tags to predict the current one. This problem greatly limits its performance, especially in identifying the Begin tags, which will be further demonstrated in Table~\ref{tbl:tag-result}.
  \item[(2)] The CRF forms strong baselines in our experiments, especially in precision. In the FoRM-M dataset, it achieves the best precision of $78.08\%$ among all the models. This is as expected, because hand-crafted local linguistic features are used in the CRF, making it easy for the model to capture the phrases with strong ``indicating words", such as ``quiz 1.1" and ``video of lecture 4". However, the recall for CRF is relatively low ($11.3\%$ lower than the proposed method in average), because in many cases, local linguistic features are not enough in identifying resource mentions, and long-range context dependencies need to be considered (e.g., the phrase ``Chain Rule" in Figure~\ref{fig:context-reference-problem}).
  \item[(3)] The BLSTM--CRF performs close to CRF on precision, but is better than CRF on recall ($+3.64\%$ in average). During prediction, the model can make use of the full context information encoded in LSTM cell rather than only local context features.
  \item[(4)] After considering character embeddings, the change of precision is not 
  obvious, but the recall improves $4.72\%$ in average compared with BLSTM--CRF. This demonstrates the effectiveness of incorporating character-level semantics. We will further analyze how character embeddings alleviates the OOV problem in Section~\ref{sec:EffectsChar}. Encoding the thread contexts further improves the recall ($+2.77\%$ in average), at the cost of a slightly drop on precision ($-1.37\%$ in average). The thread contexts bring in enough information for inferring long-term dependencies, but also burdens the model to filter out irrelevant information introduced.
  \item[(5)] As expected, the $F_1$ score of all models drops when moving from the FoRM-M to the FoRM-L dataset that contains more noisy annotations. This decrease in performance is more obvious on recall, with an average of $5.03\%$ drop. The most significant performance drop comes from CRF ($-5.9\%$ in $F_1$ score), which further exposes its limitation in handling the variability of resource mentions. The proposed method, with a $1.68\%$ decrease in $F_1$, proves to be the most robust model, owing to its high model complexity.
\end{description}

\renewcommand\arraystretch{1.2}
\begin{table}\footnotesize
	\begin{center}
		\begin{tabular}{ c|c|C{1.2cm}|C{1.2cm}|C{1.2cm}|C{1.5cm}|C{2.1cm} } \hline
            Resource Type & Tag & BLSTM & CRF & \tabincell{c}{BLSTM \\ --CRF} &  \tabincell{c}{BLSTM \\ --CRF--CE} & \tabincell{c}{BLSTM-- \\ CRF--CE--CA} \\ \hline \hline
			\multirow{2}{*}{Assessments} & B & 47.67 & 74.48 & 78.27 & 79.83 & \textbf{80.15} \\\cline{2-7}
			& I & 61.32 & 76.01 & 79.66 & 81.06 & \textbf{81.44} \\\hline
			\multirow{2}{*}{Exams}&B & 45.89 & 70.27 & 72.52 & 73.97 & \textbf{77.46} \\\cline{2-7}
			& I & 62.55 & 65.48 & 72.55 & \textbf{77.75} & 76.20 \\\hline
			\multirow{2}{*}{Videos} & B & 51.73 & 75.48 & \textbf{69.23} & 67.31 & 66.41 \\\cline{2-7}
			& I & 62.33 & 74.94 & 70.70 & \textbf{74.28} & 70.90 \\\hline
			\multirow{2}{*}{Coursewares}&B& 45.75 & 67.23 & 64.45 & 72.92 & \textbf{74.30} \\\cline{2-7}
			& I & 46.98 & 50.42 & 50.48 & 54.48 & \textbf{56.93} \\\hline
		\end{tabular}
	\end{center}
	\caption{The $F_1$ scores of different methods for each resource mention type on the FoRM-M dataset. The best results are in bold. }
	\label{tbl:tag-result}
\end{table}

To further investigate how different models perform on identifying each type of resource mention, we report models' micro-$F_1$ scores for each type of tag on the FoRM-M dataset. The results are summarized in Table~\ref{tbl:tag-result}, and we get several interesting observations.
For BLSTM, the $F_1$ score of Begin tags ($47.76\%$ in average) is much lower than that of Inside Tags ($58.29 \%$ in average). A reasonable explanation is that there are less training data for B-tags compared with I-tags, and BLSTM does not utilize the neighboring tags to predict the current one. After adding the CRF layer, the BLSTM \textendash CRF model makes a significant improvement in identifying B-tags ($+23.35 \%$ in average). Among the four mention types, the models achieve best results in identifying the Assessments. There are two reasons: (1) there are about $3$ times labeled data for the Assessments, compared with the other $3$ types, and (2) identifying the mention of assessments does not rely much on long-range contexts (e.g., ``Assignment 1.3"). The Coursewares is the most difficult resource type to identify; all models achieve the lowest $F_1$ scores in identifying the Coursewares. This is due to the high variety of this type, since it is a mixture of transcripts, readings, slides, and other additional resources. Furthermore, long-range context dependency is more common in this type (e.g., ``sgd.py"), which further increases its variety. 

\subsection{Effect of the character encoder}
\label{sec:EffectsChar}

This section examines how our introduction of the {\it Character Encoder} addresses the problem of Out-of-Vocabulary. To this end, we first evaluate the severity of the OOV problem on our data. We define OOV words as the words that cannot be found in the pre-trained GloVe embeddings, which has a vocabulary size of 400K\footnote{https://nlp.stanford.edu/projects/glove/}. As OOV words do not have pre-trained word embeddings, we need their character-level information to be taken into account. The FoRM-M dataset contains a vocabulary size of 9,761, with 3,045 (31.19\%) of them are OOV words. This reveal the severe of the OOV problem in our task. 

To understand how character encoder addresses the OOV problem, we analyze the prediction results of BLSTM--CRF and BLSTM--CRF--CE on the test set of FoRM-M, which contains 876 ground-truth resource mentions within its 839 testing examples. Among these 876 resource mentions, 163 of them contain at least one OOV word. We call these resource mentions as \textit{OOV Mentions}; identifying OOV Mentions require both word-level and character-level semantics. Other resource mentions are then denoted as \textit{None-OOV Mentions}. 

\begin{table}\footnotesize
	\begin{center}
		\begin{tabular}{ c|c|c|c } \hline
			\multirow{2}{*}{Performance}&\multicolumn{3}{c}{Correct / Total / Ratio} \\ \cline{2-4}
			&All Mentions & None-OOV Mentions & OOV Mentions \\\hline \hline
			LSTM--CRF & 564 / 876 / 64.38\% & 471 / 713 / 66.06\% &93 / 163 / 57.06\% \\\hline
		   LSTM--CRF--CE &600 / 876 / 68.49\%&493 / 713 / 69.14\%& 107 / 163 / 65.64\% \\\hline \hline
		   Improvements & 4.11\% &3.08\% & 8.58\% \\\hline
		\end{tabular}
	\end{center}
	\caption{The performance comparison between BLSTM--CRF and BLSTM--CRF--CE on the test set of FoRM-M. `Correct/Total" refers to the correct/total number of predictions, ``Ratio" is the ratio of correct prediction. }
	\label{tbl:oov-result}
\end{table}

Table~\ref{tbl:oov-result} shows the performance comparison between BLSTM--CRF and BLSTM--CRF--CE on both the OOV mentions and none-OOV mentions. Among the 876 testing resource mentions, the rate of correct predictions\footnote{A correct prediction means that the prediction of scope and type for a resource mention are both correct. } increases from $64.38\%$ to $68.49\%$, with a $4.11\%$ improvement. But the performance improvement for the none-OOV mentions only increase $3.08\%$. For the OOV mentions, however, the performance boost is $8.58\%$, much higher than the overall improvement of performance. This indicates that incorporating character-level information significantly benefits the identification of OOV resource mentions, which makes a major contribution to the overall performance improvement. 

\subsection{Error Analysis}
\label{sec:ErrorAnalysis}

\renewcommand\arraystretch{1.3}
\begin{table}\scriptsize
	\begin{center}
		\begin{tabular}{ c | c } \hline
			Types & Examples \\\hline \hline
			Exactly & (Pred.) Problem with fminunc when I run $\textbf{[ex2regm]}_{Coursewares}$.\\
			Correct &(G.T.) Problem with fminunc when I run $\textbf{[ex2regm]}_{Coursewares}$. \\\hline
			\multirow{2}{*}{Missing} &(Pred.) I was wondering about the file location of house\_data\_g1. \\
			& (G.T.) I was wondering about the file location of $[\textbf{house\_data\_g1]}_{Coursewares}$. \\ \hline
			Wrongly & (Pred.) I was wondering about $\textbf{[the file]}_{Coursewares}$ location of house\_data\_g1.\\
			Extracted & (G.T.) I was wondering about the file location of house\_data\_g1. \\\hline
            Scope Wrong & (Pred.) Hi, I completed $\textbf{[the quiz]}_{Exams}$ for week 2. \\
			Type Right & (G.T.) Hi, I completed $\textbf{[the quiz for week 2]}_{Exams}$. \\\hline
			Scope Right &(Pred.) How about to understand it better from$\textbf{[the simulation lecture]}_{Assessments}$. \\
			Type Wrong &(G.T.) How about to understand it better from $\textbf{[the simulation lecture]}_{Videos}$.
			 \\\hline
			Scope Wrong & (Pred.) Anybody have the zip file of $\textbf{[assignment]}_{Assessments}$ for week 3?\\
			Type Wrong & (G.T.) Anybody have $\textbf{[the zip file of assignment for week 3]}_{Coursewares}$? \\\hline
		\end{tabular}
	\end{center}
	\caption{Prediction error types and examples. (Pred.) is the model prediction, and (G.T.) is the ground truth. The bold texts with $[\:\:\:]$ are identified/true reource mentions associated with type label. }
	\label{tbl:error-examples}
\end{table}

The micro-$F_1$ score is a proper evaluation metric for models' performance on individual tags; however, does not tell us why errors are made.
To provide an in-depth analysis of the proposed model's performance, we list the six possible conditions that happen during the prediction, summarized in Table~\ref{tbl:error-examples}, together with examples. The model makes an \emph{Exactly Correct} prediction if the scope of the prediction exactly matches the ground truth, and the predicted type is also correct. There are cases when scopes are matched but the predicted type is incorrect or conversely, these are summarized as three cases: \emph{Scope Right/Wrong Type Right/Wrong}. The remaining conditions happen when the prediction has no overlap with the ground truth in the sentence, which are divided into \emph{Missing} and \emph{Wrongly Extracted} errors.

\begin{table}\scriptsize
	\begin{center}
        \renewcommand\arraystretch{1.4}
		\begin{tabular}{ m{2cm}<{\centering} |c|c|c|c|c } \hline
		Types	& Assessments & Exams & Videos & Coursewares & Total \\\hline \hline
		Exactly Correct & 385 (69.4\%)& 61 (68.6\%)& 66 (54.1\%) & 88 (59.5\%) & 600 (65.6\%)\\\hline
		Missing & 15 (2.7\%) & 2 (2.2\%) & 4 (3.3\%) & 19 (12.8\%) & 40 (4.4\%) \\\hline
		Wrongly Extracted & 19 (3.4\%) & 3 (3.4\%)& 5 (4.1\%)& 11 (7.4\%)& 38 (4.2\%)\\\hline
		Scope Wrong Type Right & 129 (23.2\%) &  22 (24.7\%) & 44 (36.1\%) & 20 (13.5\%) & 215 (23.5\%)  \\\hline
		Scope Right Type Wrong &4 (0.7\%)& 1 (1.1\%)& 1 (0.8\%) & 4 (2.7\%)& 10 (1.1\%) \\\hline
		Scope Wrong Type Wrong &3 (0.6\%)& 0 (0.0\%)& 2 (1.6\%)& 6 (4.1\%)& 11 (1.2\%) \\\hline \hline
		Total & 555 & 89 & 122 & 148 & 914 \\\hline 		
		\end{tabular}
	\end{center}
	\caption{Error analysis of BLSTM--CRF--CE--CA on the FoRM-M dataset. The table shows the number/precetage of different prediction cases for different resource types. }
	\label{tbl:error-types}
\end{table}

Table~\ref{tbl:error-types} summarizes the performance of BLSTM--CRF--CE--Context on the FoRM-M test set. Among all the $914$ cases obtained from the $839$ testing examples, $600 (65.6\%)$ of them are predicted completely correctly by the model. We observe that most of the errors come from the \emph{Scope Wrong Type Right}, holding a high percentage of $23.5\%$. Compared to this, other errors are less obvious. However, we further discover that a large portion ($178$ out of $215$ cases) of this error happens because the model selects a more `general" mention from a longer ground truth. For example, as given by the example in Table~\ref{tbl:error-examples}, the model selects the phrase ``the quiz" from the ground truth mention ``the quiz for week 2". This behavior can be explained by the feature of sequence tagging; the decoder tends to select shorter and general patterns, as they are more frequently present as training signals. To some extent, both general and specific mentions are acceptable in practice, but teaching model to identify more specific mentions is a future direction for improvement. A potential solution is to take into account the grammatical structure of the sentence in decoding. Another observation is that besides the scope error, the \emph{Missing} error holds a high percentage of $12.8\%$ in identifying Coursewares. This is consistent with the relative low recall presented in Table~\ref{tbl:final-result}, which poses the challenges of dealing with noisy expressions and long-range context dependency. Encoding thread context partially addresses the challenge, but there is still much room for improvement. 

\section{Conclusion and Future Works}
\label{sec:conclusion}
We propose and investigate the problem of automatically identifying resource mentions in MOOC discussion forums. 
We precisely define the problem and introduce the major challenges: the variety of expressions and the context dependency. 
Based on the vanilla LSTM--CRF model, we propose a character encoder to address the OOV problem caused by the variety of expressions, and a context encoder to capture the information of thread contexts. 
To evaluate the proposed model, we manually construct a large scale dataset FoRM based on real online forum data collected from Coursera. The FoRM dataset will be published as the first benchmark dataset for this task. Experimental results on the FoRM dataset validate the effectiveness of the proposed method. 

To build up a more efficient and interactive environment for learning and discussing in MOOCs, it requires the interlinkings between resource mentions and real resources. Our work takes us closer towards this goal. A promising future direction is to investigate how to properly resolve the identified resource mentions to real learning resources. However, it is also worthy to notice that the current identification performance still has much room for improvement; there are still challenges that are not fully addressed, such as identifying more specific resource mentions, as discussed in Section~\ref{sec:ErrorAnalysis}. Addressing these challenges by utilizing more features from both static materials and dynamic interactions in MOOCs are also promising future directions to be explored.

\section*{Acknowledgements}
This research is funded in part by a scholarship from the
China Scholarship Council (CSC), a Learning Innovation -- Technology grant from NUS (LIF-T, NUS) and UESTC Fundamental Research Funds for the Central Universities under Grant No.: ZYGX2016J196.

\section*{References}
\bibliography{mybibfile}

\newpage
\appendix
\section{Annotation Details}
\label{sec:appendix}
We train the annotators in advance, before starting the annotation at June, 2018. First, we email every annotator with an annotation instruction document, which contains the detailed description and examples for different types of resource resources, {\it cf} Table~\ref{table:tab-types-resource}. We then provide them with a link to our brat platform with an example annotation file containing formatted annotation data and typical examples. They are required to complete an one hour training to learn the usage of the annotation tool and try out some practical annotations to better understand the annotation instruction. Finally, we add every annotator to a Wechat group to coordinate questions and answers about unclear examples. We observe that a few questions are raised at the beginning of the annotation, and later the annotators become more confident and fluent in their annotation. 

\begin{table}\scriptsize
	\begin{center}
		\begin{tabular}{ c | c | l } \hline
Types & Description & Examples \\ \hline \hline
\multirow{4}{*}{Assessments} & \multirow{4}{6cm}{any form of assessments, exercises, homeworks and assignments.} & the/that/this/my assignment;\\
            && assignment 3;\\
            &&question 1 in assignment 2;\\
            &&assignment [name].\\ \hline
\multirow{4}{*}{Exams} & \multirow{4}{6cm}{evaluate the knowledge and/or skills of students, including quizzes, tests, mid-exams and final exams.} & the/that/this quiz;\\
            &&quiz 3;\\
            &&question 1 in quiz 2;\\
            &&quiz [name]. \\ \hline
\multirow{4}{*}{Videos} & \multirow{4}{6cm}{videos present the lecture content.} & this/that/the video;\\
&&this/that/the lecture;\\
&&video [name];\\
&&slide 4 in video. \\ \hline

\multirow{3}{*}{Readings} & \multirow{3}{6cm}{optionally provide a list of other learning resources provided by courses.} & the exercise instructions; \\
            &&this week's video tutorials;\\
            &&the lecture notes.\\ \hline
\multirow{3}{*}{Slides} & \multirow{3}{6cm}{slides provide the lecture content for download and separate review, often aligned to those in the video.} & slide pdfs;\\
&&slide 2;\\
&&the lecture slides.\\ \hline
\multirow{2}{*}{Transcripts} & \multirow{2}{6cm}{transcripts or subtitles of the videos.} & transcript of video 1;\\
&&the subtitle files. \\ \hline
\multirow{2}{*}{Additional} & \multirow{2}{6cm}{help to catch other materials made available for specialized discipline-specific courses.} &ex2regm.m;\\
Resources&&ex2reg scripts.\\
&& \\ \hline
		\end{tabular}
	\end{center}
\caption{The description and examples for the pre-defined $7$ types of resources mentions.}
\label{table:tab-types-resource}
\end{table}

\end{document}